\documentclass[lettersize,journal]{IEEEtran}
\usepackage{amsmath,amsfonts}
\usepackage{algorithmic}
\usepackage{algorithm}
\usepackage{array}
\usepackage[caption=false,font=normalsize,labelfont=sf,textfont=sf]{subfig}
\usepackage{textcomp}
\usepackage{stfloats}
\usepackage{url}
\usepackage{verbatim}
\usepackage{graphicx}
\usepackage{cite}
\usepackage{multirow}
\usepackage[flushleft]{threeparttable}

\begin{document}

\title{Object State Estimation Through Robotic Active Interaction for Biological Autonomous Drilling}

\author{Xiaofeng Lin, Enduo Zhao, Saúl Alexis Heredia Pérez and Kanako Harada, \IEEEmembership{Member,~IEEE}
\thanks{Research supported in part by JST Moonshot R\&D JPMJMS2033 and in part by the University of Tokyo Fellowship. \emph{(Corresponding authors: Xiaofeng Lin.)}}
\thanks{The first and second authors contribute equally to this research.}
\thanks{Xiaofeng Lin is with the Graduate Schools of Medicine \& Engineering, the University of Tokyo. Enduo Zhao is with the Department of Mechanical Engineering, the University of Tokyo. Saúl Alexis Heredia Pérez, and Kanako Harada are with the Center for Disease Biology and Integrative Medicine, Graduate School of Medicine, The University of Tokyo. }}

\markboth{Journal of \LaTeX\ Class Files,~Vol.~14, No.~8, August~2021}%
{Shell \MakeLowercase{\textit{et al.}}: A Sample Article Using IEEEtran.cls for IEEE Journals}


\maketitle

\begin{abstract}

Estimating the state of biological specimens is challenging due to limited observation through microscopic vision. For instance, during mouse skull drilling, the appearance alters little when thinning bone tissue because of its semi-transparent property and the high-magnification microscopic vision. To obtain the object's state, we introduce an object state estimation method for biological specimens through active interaction based on the deflection. The method is integrated to enhance the autonomous drilling system developed in our previous work. The method and integrated system were evaluated through 12 autonomous eggshell drilling experiment trials. The results show that the system achieved a 91.7\% successful ratio and 75\% detachable ratio, showcasing its potential applicability in more complex surgical procedures such as mouse skull craniotomy. This research paves the way for further development of autonomous robotic systems capable of estimating the object's state through active interaction.

\end{abstract}


\section{Introduction}
\label{sec:introduction}

\begin{figure}[!t]
\centering
\includegraphics[width=2.4in]{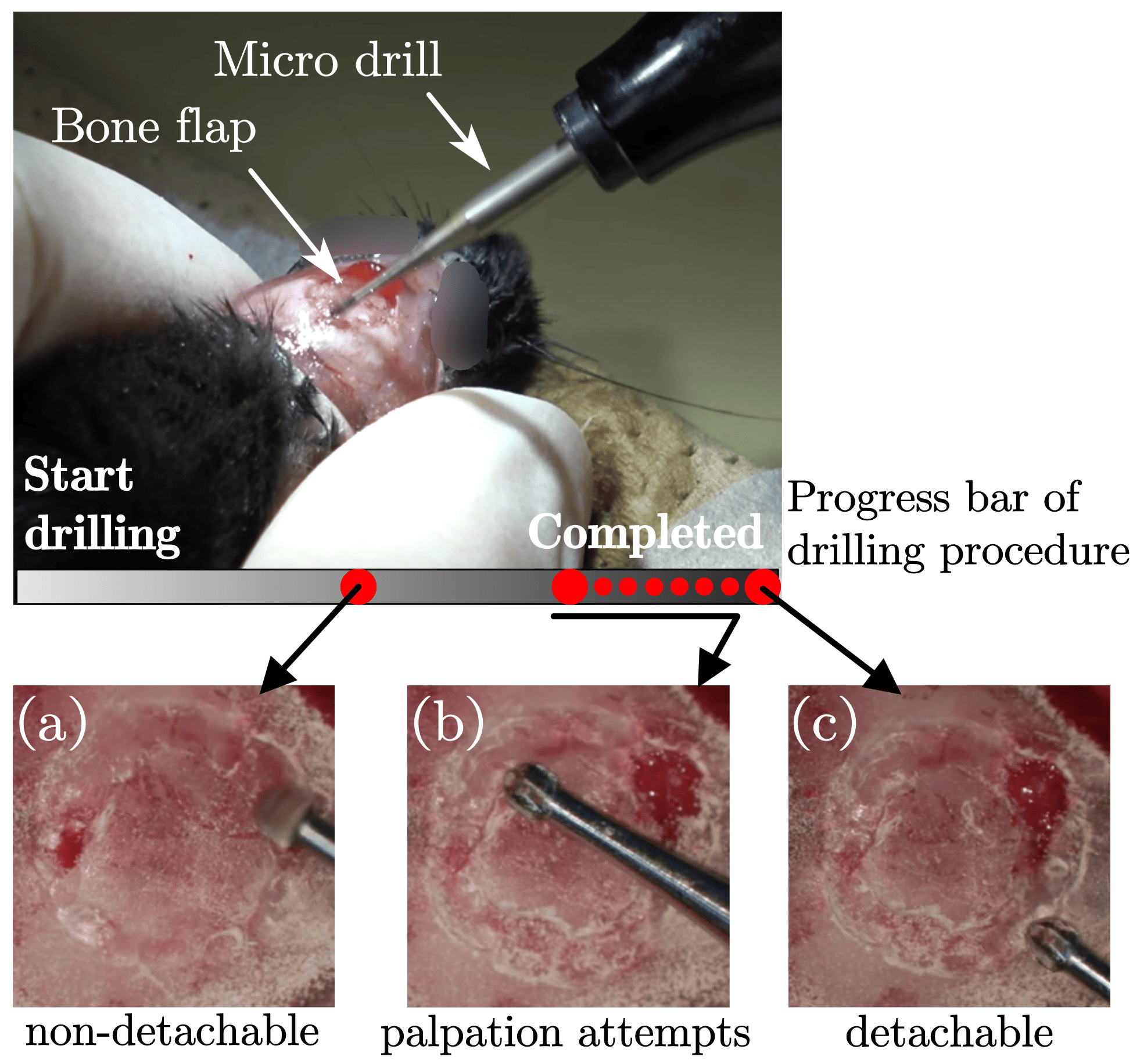}
\caption{Demonstrations of cranial window creation on mice skull (performed by surgeons in Institute of Science Tokyo). (a) The state of bone flap was still non-detachable on the way of drilling. (b) The state of bone flap was about to change and the operator started palpating. (c) Finally, the state was confirmed to be detachable.}
\label{fig:intro}
\end{figure}

Object state estimation of biological specimens during micromanipulation presents a significant challenge for human operators not only due to the properties of the specimen itself, such as a semi-transparent property or being encased in a viscous liquid, but also due to the limitations in the high magnification observation, where visual cues are insufficient for indicating the minute change. For example, in scientific experiments involving the transplantation of human organoids into mice and observing their growth \cite{takebe2019}, a crucial preliminary task is to create an 8-mm cranial window on the mouse skull using a handheld microdrill under a high-magnification microscope. During this process, operators find it difficult to determine the completion of the drilling task by estimating the state of bone flap through visual inspection alone, which is defined in two cases: non-detachable and detachable. A non-detachable state means that the bone flap is drilled but not yet isolated from the body (see Fig. \ref{fig:intro}-(a)), while a detachable state means that the bone flap is isolated from the other parts and can be removed manually (see Fig. \ref{fig:intro}-(c)).

Many studies have focused on developing robotic systems to achieve semi-automation or automation of the cranial window creation process, where the estimation of bone flap state has always been a challenge. Ghanbari et al. \cite{ghanbari2019} developed the “Craniobot”, an cranial microsurgery platform that uses micro-CT scans to measure skull thickness and calculate target drilling depth, assuming the bone flap becomes detachable once the target depth was reached. However, uneven skull thickness and individual differences in mice make accurate pre-measurement impractical. Pak et al. \cite{pak2015} discovered that a sudden increase in electrical conductance between the drill bit and the mouse skull occurs when the drill bit penetrates the skull without contacting the brain. Based on this observation, they developed a conductance-based algorithm to monitor the completion of drilling and implemented a robotic system for automated craniotomies.

In our previous work, the Autonomous Robotic Drilling System (ARDS) \cite{zhao2023} was developed to automate the drilling process and evaluated using eggshells due to their similar physical properties to the cranial shell. This innovative system includes a trajectory planner that generates and updates the drilling trajectory, and a 2D-image-based completion level recognition block processing images captured by a 4K camera positioned above the operation area using a neural network to recognize the drilling completion level of sample points along the trajectory. The state estimation of the bone flap was accomplished by averaging the drilling completion levels of all sample points, assuming that an initially non-detachable state would become detachable once the average completion level reached 100\%. However, due to the limitation of the recognition accuracy on 2D images, the state estimation of bone flap was sometimes imprecise, particularly when the average completion level was over 80\%. Therefore, only 10 out of 20 trials on eggshell drilling were actually detachable when the state of the bone flap was estimated as detachable in the previous experiments. In 6 of the remaining 10 non-detachable trials, bone flaps could be removed \textbf{forcibly} with tweezers by humans without damaging the membrane beneath, which were still considered successful trials and led to an 80\% of drilling successful ratio but involving forcible removal reduced the objectivity of the results and posed a risk of damaging the underlying membrane. 

The proposed methods in the aforementioned studies indirectly estimate the state of the bone flap by processing signals or information generated during the drilling procedure, but these methods have certain accuracy issues. Combining the results of our previous experiments, it can be concluded that object state estimation methods based on indirect signals are unreliable. As a result, an accurate, reliable, and objective state estimation method for bone flaps without pre-measurement remains a significant challenge. 

\begin{figure*}[!t]
\centering
\includegraphics[width=7in]{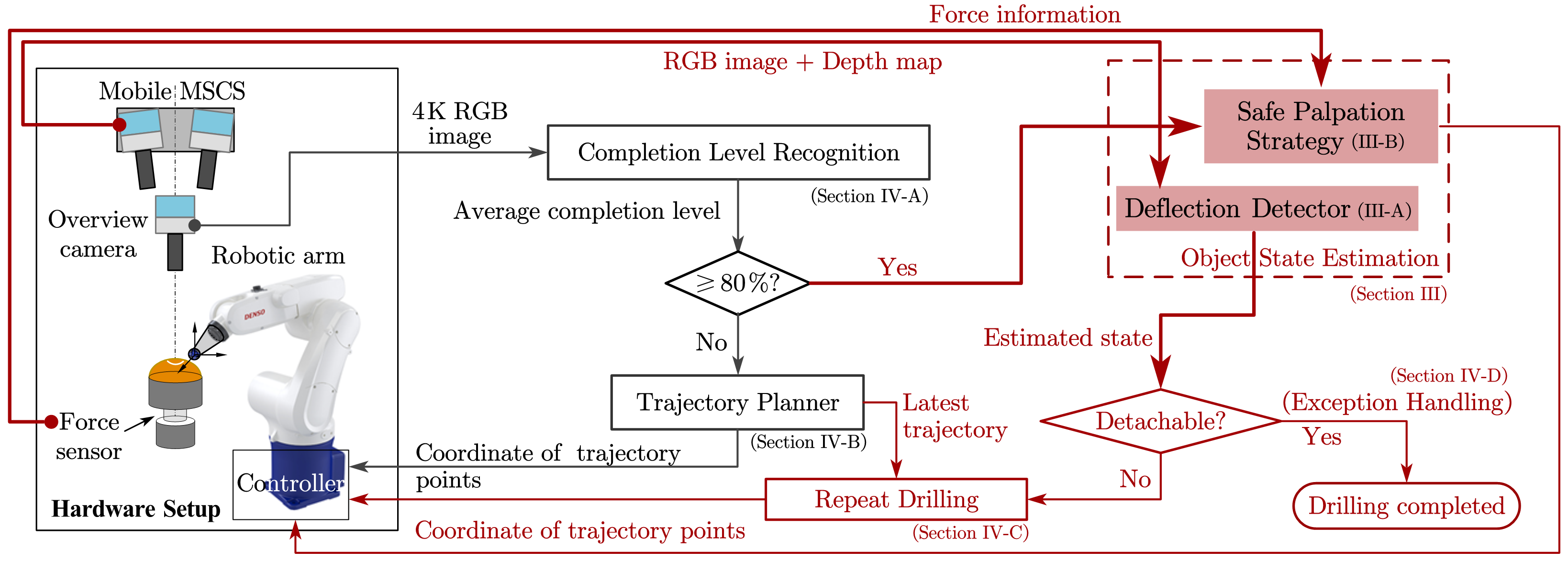}
\caption{Overview of the improved ARDS. The parts with black text and borders are the same as the original system \cite{zhao2023}, while the parts with red text and borders are newly added in this work.}
\label{fig:overview}
\end{figure*}

Inspired by human operators estimating the state of the bone flap by palpating it with a drill bit in practical mouse skull drilling (see Fig. \ref{fig:intro}-(b)), actively interacting with the specimen to estimate its state is considered a feasible approach as it generates detectable signal changes. Hasegawa et al. \cite{hasegawa2023} proposed an active interaction method by dividing the drilling trajectory into 32 sections and pressing each section to obtain and process force signals for penetration detection of the section. They predefined the robotic motion and monitored the force signal during the palpation, which is time-consuming because the method can only estimate the state in small areas with one palpation trial. Besides, they checked the state only at the last phase of the procedure, lacking real-time capability. Therefore, We aim to propose a direct method that processes global signals for object state estimation.


In this study, the ARDS proposed in our preliminary research \cite{zhao2023} was further improved by introducing an active interaction-based object state estimation method for the biological specimen drilling task. The core of the method is that in the behavior of palpation, the operator assumes that if an observable deflection occurs in the direction of pressing, it indicates that the bone flap is detachable; otherwise, the bone flap is non-detachable and drilling is continued. However, relying solely on the 2D images obtained from the original 4K camera in the ARDS is still insufficient to observe the deflection brought about by the active interaction, and the commercially available 3D camera solutions, which are usually used to detect deflections, can not meet our requirements in accuracy. Therefore, we adopted the Microscopic Stereo Camera System (MSCS) \cite{lin2024} that offers a precision of 0.10 ± 0.02 mm at a frequency of up to 30 Hz, which can detect minute deflections of the bone flap, ranging from 0.05 mm to 0.30 mm (based on the experiments). The object state estimation method is integrated into the ARDS to improve the automation workflow, achieving an increased detachable ratio and a higher successful drilling ratio in autonomous eggshell drilling experiments. The main contributions of this work can be summarized as follows:

\begin{enumerate}
    \item An object state estimation method via active robotic interaction was proposed.
    \item The method was integrated with the ARDS and the autonomous drilling workflow was optimized.
    \item The effectiveness and performance of the improved system was validated through 12 trials of eggshell autonomous drilling experiments.
\end{enumerate}

\section{System overview and hardware setup}\label{sec:systemsetup}

\begin{figure}[!t]
\centering
\includegraphics[width=2.4in]{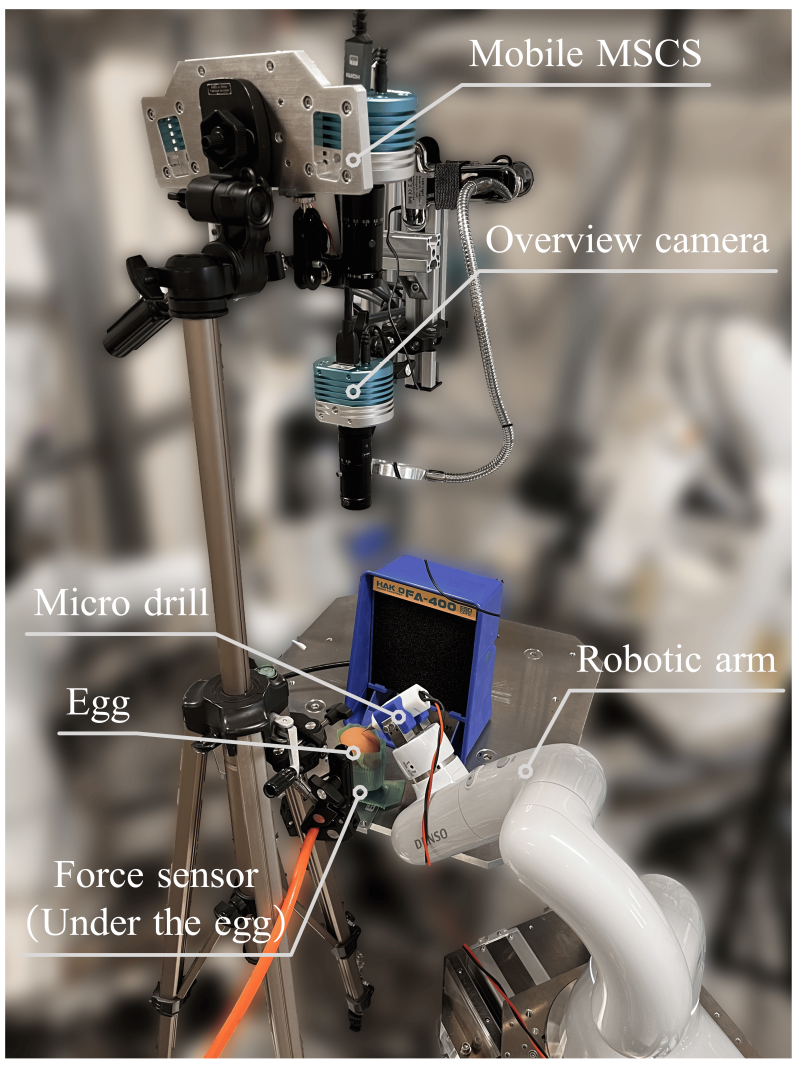}
\caption{The hardware setup of the ARDS consists of a robotic arm holding a micro drill, an egg to be drilled, a force sensor, an overview camera, and a mobile MSCS.}
\label{fig:setup}
\end{figure}

The overview of the ARDS improved by the proposed object estimation method is shown in Fig.~\ref{fig:overview}. Object State Estimation module, which is one of the main contributions of this study, will be introduced in Section~\ref{subsec:displacementdetector}. The autonomous drilling workflow is further optimized by integrating it with other modules (Completion Level Recognition, Trajectory Planner, and Repeat Drilling). Other modules, as well as the entire workflow, will be introduced in Section~\ref{subsec:automationflow}.

The hardware setup of the system is shown in Fig.~\ref{fig:setup}, utilizing one of the robotic arms from the Multiarm Robotic Platform for scientific exploration \cite{marinho2024}. The robotic arm, denoted as $\mathbb{R}$, is an 8-degree-of-freedom serial manipulator with joint values $q \in \mathbb{R}^{8}$, consisting of the CVR038 robotic arm (Densowave, Japan), a linear actuator, and a circular rail actuator. The arm holds a micro drill (MD1200, Braintree Scientific, USA) to drill objects fixed by a clamping mechanism, which is equipped with a force sensor (Thin NANO sensor, BL AUTOTECH Ltd, Japan) at the bottom. An overview camera (STC-HD853HDMI, Omron-Sentech, Japan) with a low-distortion macro lens (VS-LDA75, VS Technology, Japan) is positioned above to capture 4K images for the Completion Level Recognition module. Due to the different image resolution and frame rate requirements for Object State Estimation ($540 \times 960$, 20 Hz) and Completion Level Recognition (4K, 30 Hz), we developed a mobile MSCS setup mounted on a tripod, based on the original desktop MSCS setup detailed in \cite{lin2024}, to adapt to the constrained workspace of the robotic system room. The cameras and lens of mobile MSCS are the same as the overview camera for the stereo configuration. The reconstruction space of the stereo vision is approximately $20 \times 30 \times 10 \ mm$ and the working distance is around $500 \ mm$. Additionally, the mobile MSCS also incorporates a dual laser module (VLM-650-03, Quarton, Taiwan) to enable the fast and rough positioning of the microscopic 3D space despite the focus.

\section{Object State Estimation}
\label{subsec:displacementdetector}

The proposed object state estimation method aims to detect the deflection of the bone flap through active interaction to estimate its state. The method contains two parts: the Defection Detector to monitor the deflection of the bone flap with the input of RGB-Depth (RGB-D) images (Section~\ref{subsec:deflection}), and the Safe Palpation Strategy to control the drill bit to press the bone flap safely with the input of contact force (Section~\ref{subsec:palpation}).

\subsection{Deflection Detector}\label{subsec:deflection}

With the RGB-D images obtained by MSCS input, the deflection of the bone flap can be detected and the two states (non-detachable and detachable) can be defined by the Deflection Detector, shown in Fig.~\ref{fig:system}. The components of the Deflection Detector are described as follows:

\begin{figure}[!t]
\centering
\includegraphics[width=3in]{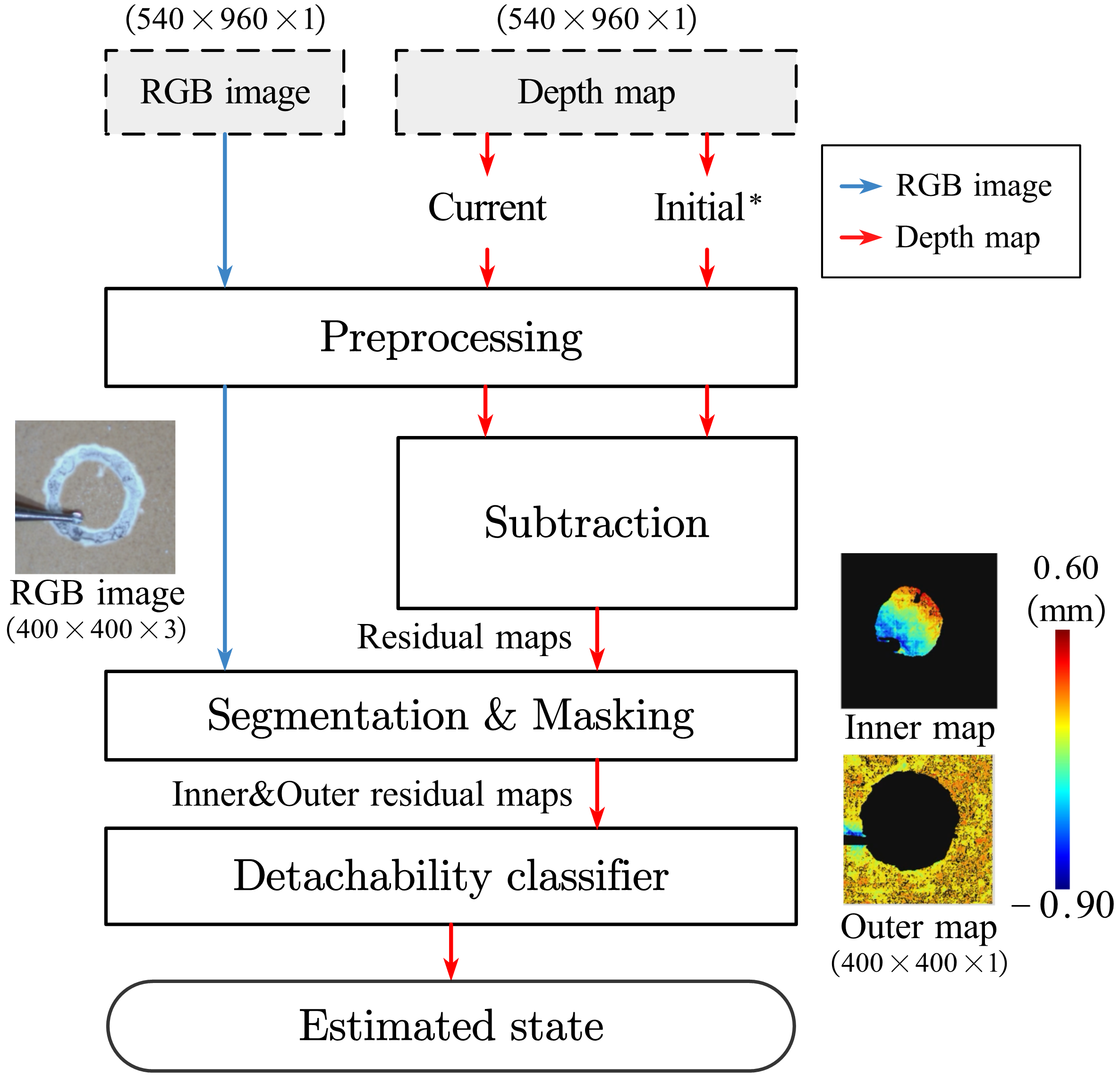}
\caption{The workflow of the Deflection Detector. It discriminates the inner and outer areas from RGB images and calculates the deflection of the two areas in pixels by their disparity values through the Detachability classifier. The state is estimated based on the relative mean depth values of the two areas. *The input involves an initial depth map before the state change.}
\label{fig:system}
\end{figure}

\subsubsection{Preprocessing}
As the interaction will result in a change in depth, the comparison should be made between the initial and current depth maps. In this paper, the egg was well-fixed and assumed to be static during the drilling. Thus, the initial depth map is captured before the drilling. The RGB images and depth maps are captured from the MSCS, where the RGB images are used to identify and segment inner and outer areas in the Segmentation \& Masking. The Preprocessing module includes a cropping operation, where a fixed central area of the image containing the drilling path is cropped to enable faster processing.  

\subsubsection{Subtraction}
This module compares the depth maps based on the principles mentioned above. Subtraction calculates the residual map of the initial and current depth maps, indicating the overall change in depth caused by the interaction. 

\subsubsection{Segmentation \& Masking}
The depth map comparison focuses on regions of interest on the target surface, where in this research, the regions are the inner and outer areas of the bone flap, which can be distinguished from its RGB appearance. For example, the segmentation classifies the flap area (inner area) and the area outside of the drilling path (outer area) from the RGB image and outputs their masks, indicating different regions. The pixels of the masks are assigned unique labels to identify the regions. The research demonstrates the segmentation utilizing Hue-Saturation-Value (HSV) color thresholding for eggshell drilling operations. The labeling is implemented through the Watershed Method\footnote{https://en.wikipedia.org/wiki/Watershed\_(image\_processing)}, a typical image processing technique implemented in OpenCV software. Then, in the Masking step, the overall residual map is segmented into two sub-maps using the labeled masks from the segmentation result. In short, through this module, the overall residual map indicating changes in depth caused by the interaction is segmented into two sub-maps regarding the region of interest. 

\subsubsection{Detachability classifier}
The state of the bone flap is classified using the residual depth sub-maps. In the case of the drilling experiment, the sub-maps indicate the flap area and body area, respectively. The classifier compares the mean depth values $\overline{D_i}, \overline{D_o}$ of the pixels in the inner and outer maps. Concerning the detachable state of the bone flap, the difference between mean depth values $d\overline{D} = \overline{D_i}-\overline{D_o}$ is used to determine whether the bone flap has been isolated from the eggshell body. If $|d\overline{D}|$ is larger than a designated threshold $|d\overline{D}|_{max}$, it will output a detachable state, or it will output a non-detachable state. Specifically, the threshold is set empirically equivalent to the average thickness of the specimen (eggshell). 

\subsection{Safe Palpation Strategy}\label{subsec:palpation}

Safe Palpation Strategy involves controlling the drill bit to contact the bone flap while monitoring the contact force to ensure it stays below a safety threshold. 

Assuming uneven drilling path completion, the system touches 4 strategic points along the bone flap edge to maximize the likelihood of creating deflection. The selection of the positions of these 4 strategic points is detailed in the supplementary document. As for one strategic point, the drill bit descends at a controlled velocity $v_z$, with the estimated state output from the Deflection Detector (described in Section~\ref{subsec:deflection}) and the contact force $F_z$ output from the force sensor monitored. If the estimated state becomes detachable while the contact force is below the maximum safe value $|F_z|_{max}$, the strategic point is considered detachable; if the estimated state remains non-detachable until the contact force exceeds the safe value, the drill bit descent is immediately halted, and the strategic point is considered non-detachable. If 3 out of 4 strategic points are detected as detachable, the state of the whole bone flap is considered detachable, completing the drilling.



\section{Autonomous drilling workflow}\label{subsec:automationflow}

With the integration of the proposed object state estimation method (Section~\ref{subsec:displacementdetector}), an improved ARDS is developed. In this section, we introduce the optimized autonomous drilling workflow (see Fig.~\ref{fig:overview}) and other essential modules of the system (Section~\ref{subsec:completionrecognition}, Section~\ref{subsec:trajectoryplanner} and Section~\ref{subsec:repeatdrilling}). Additionally, the system is now better equipped to handle exceptions that may arise during the drilling procedure (Section~\ref{subsec:exception}).

\subsection{Completion Level Recognition}\label{subsec:completionrecognition}

This module is consistent with our original system \cite{zhao2023}. In the optimized workflow, the Completion Level Recognition module uses a neural network to recognize and update the completion levels of 32 sample points along the drilling path based on 4K RGB images from the overview camera, and then output the average completion level of these points. As using the average completion level to estimate the state of the bone flap becomes inaccurate when it exceeds 80\% (as discussed in Section~\ref{sec:introduction}), we devised a strategy: drilling continues if the average completion level is below 80\%, but pauses when it exceeds 80\%, and the system transitions to the Object State Estimation module (Section~\ref{subsec:displacementdetector}). If the estimated state output from the Object State Estimation is detachable, drilling stops; if non-detachable, Repeat Drilling is triggered (Section~\ref{subsec:repeatdrilling}).

\subsection{Trajectory Planner}\label{subsec:trajectoryplanner}

This module is the same as our original system \cite{zhao2023}, which outputs the coordinates of points on a continuous drilling trajectory based on the input of recognized completion levels of 32 discrete sample points along the drilling path. The module includes a multimodal velocity damper to control the descent speed of the drill bit, allowing it to adapt to different completion levels at different sample points and update their z-axis coordinates, and a continuous trajectory generation module, which generates a continuous trajectory by using the coordinates of the discrete sample points through a constrained cubic spline interpolation method.

\subsection{Repeat Drilling}\label{subsec:repeatdrilling}

During the Repeat Drilling process, the Completion Level Recognition is temporarily suspended, allowing the drilling along the latest trajectory to be executed for an additional ten cycles without any modifications. This is based on the assumption that when the average completion level reaches 80\%, the latest trajectory is sufficient to make the bone flap detachable, but some minor adjustments are still necessary to transition from a non-detachable to a detachable condition. The focus of this process is on removing residual bone along the same trajectory, which may have been preserved due to the inherent elasticity of the bone. Upon concluding the Repeat Drilling, the system resumes the Object State Estimation process. This cycle continues until the bone flap is assessed to be detachable, at which point the drilling process is deemed complete.

\subsection{Exception Handling}\label{subsec:exception}

During drilling, the eggshell may crack or break due to prolonged contact with the drill bit, which our previous 2D image-based system \cite{zhao2023} could not handle autonomously. Manual intervention was required to stop drilling and verify results. By introducing the Deflection Detector (see~\ref{subsec:deflection}), the system can identify such anomalies by monitoring the deflection of the bone flap in real-time. Therefore, we keep the Deflection Detector operational throughout the entire drilling process in addition to the Object State Estimation phase. If significant deflection is observed during the drilling process, the system will stop the drilling operation to allow for manual safety checks.

\section{Experiments}

In this section, we conducted 12 trials of eggshell drilling experiments to evaluate the effectiveness and performance of the ARDS improved by the proposed object state estimation method. It is important to note that the effectiveness of the proposed object state estimation method has been thoroughly evaluated in the supplementary document. Therefore, in the evaluation of the improved ARDS in this section, the method is considered to be completely accurate as part of the system. The improved ARDS is shown in Fig.\ref{fig:setup}, and the hardware setup is described in Section\ref{sec:systemsetup}.

\subsection{System configuration and setup}

The experiments were conducted on an Ubuntu 20.04 x64 system with a software implementation. The robotic arm was controlled according to the method described in \cite{marinho2024}. For interprocess communication, ROS (Robot Operating System) Noetic Ninjemys was employed, and simulations were performed using CoppeliaSim (Coppelia Robotics, Switzerland). Communication with the robot was facilitated through the SmartArmStack\footnote{https://github.com/SmartArmStack}. The dual quaternion algebra and robot kinematics were implemented using DQ Robotics \cite{adorno2021dqrobotics} with Python3. A Raspberry Pi 4 Model B was used to interface with the force sensor, and the force data was collected at 128 samples per second (SPS) using Python’s SMBus module along with an I2C analog input board.

We set the resolution and frame rate of the images to be $4K, \ 30 \ Hz$ the overview camera and $540 \times960, \ 20 \ Hz$ for the mobile MSCS to meet the different requirement of the Completion Level Recognition and Object State Estimation. 

The parameters chosen for the experiments (described in Section~ \ref{subsec:displacementdetector}), namely the force threshold $|F_z|_{max}$,  the depth threshold $|d\overline{D}|_{max}$, the decent speed $v_{z}$ of the drill, and the radius of the circular drilling path $r$ are selected to be $|F_z|_{max} = 0.40 \ N$, $|d\overline{D}|_{max} = 0.12 \ mm$, $v_{z} = -0.05 \ mm/s$ and $r = 4 \ mm$. Additionally, HSV thresholds are chosen properly as per the eggs for segmentation performance. 

\subsection{Evaluation metrics}


In our previous experiments \cite{zhao2023}, ARDS performance was evaluated using success metrics, where an experiment was considered successful if the bone flap could be manually removed without damaging the membrane. However, as mentioned in Section~\ref{sec:introduction}, this metric involves forcible removal, reducing objectivity. Therefore, we also introduced detachability metrics, independent of success metrics, based on whether the bone flap becomes detachable after drilling. Additionally, the total drilling time and the palpation time within it are recorded to analyze the system's efficiency and the impact of the added palpation stage on the system.




\subsection{Results and discussions}

The results of all 12 trials are shown in Table~\ref{table:1}. In the 12 experimental trials, the successful ratio of 91.7\% (previous system: 80\%). Meanwhile, in 9 trials, the bone flaps were judged as detachable, resulting in a detachable ratio of 75\% (previous system: 50\%). The average total drilling time of the 12 trials is 2052 seconds, which is longer than the previous system's (1008 seconds).

\begin{table}[!t]
\begin{center}
\caption{Results of 12 autonomous drilling trials on eggs}\label{table:1}
\begin{tabular}{lcccccc}
\hline
\multirow{2}{*}{No.}    &\multirow{2}{*}{Successful}    &\multirow{2}{*}{Detachable} & \multicolumn{2}{c}{Time(s)} \\ 
    &             &                                          &Total               &Palpation \\\hline
1   & ×                          &\checkmark     & 750     & 148 (19.7\%)                        \\
2   & \checkmark                 &\checkmark     & 1152    & 370 (32.1\%)                        \\
3   & \checkmark                 &\checkmark     & 906     & 296 (32.7\%)                        \\
4   & \checkmark                 &\checkmark     & 653     & 222 (34.0\%)                        \\
5   & \checkmark                 &\checkmark     & 2243    & 962 (42.9\%)                        \\
6   & \checkmark                 &×              & 1632    & 517 (31.7\%)                        \\
7   & \checkmark                 &\checkmark     & 272     & 0   (0\%)                           \\
8   & \checkmark                 &\checkmark     & 1600    & 518 (32.4\%)                        \\
9   & \checkmark                 &\checkmark     & 1230    & 221 (18.0\%)                        \\
10  & \checkmark                 &\checkmark     & 2032    & 518 (25.5\%)                        \\
11  & \checkmark                 &×              & 2355    & 963 (40.9\%)                        \\
12  & \checkmark                 &×              & 1764    & 519 (29.4\%)                        \\ \hline
Avg.& 91.7\%                     &75\%           & 2052    & 577 (28.1\%)                 \\ \hline
\end{tabular}
\end{center}
\end{table}

Based on the results, it can be concluded that the proposed object state estimation method effectively improves the performance of the ARDS in eggshell drilling, both in terms of success and detachable ratio. 

\begin{figure}[!ht]
\centering
\includegraphics[width=3.45in]{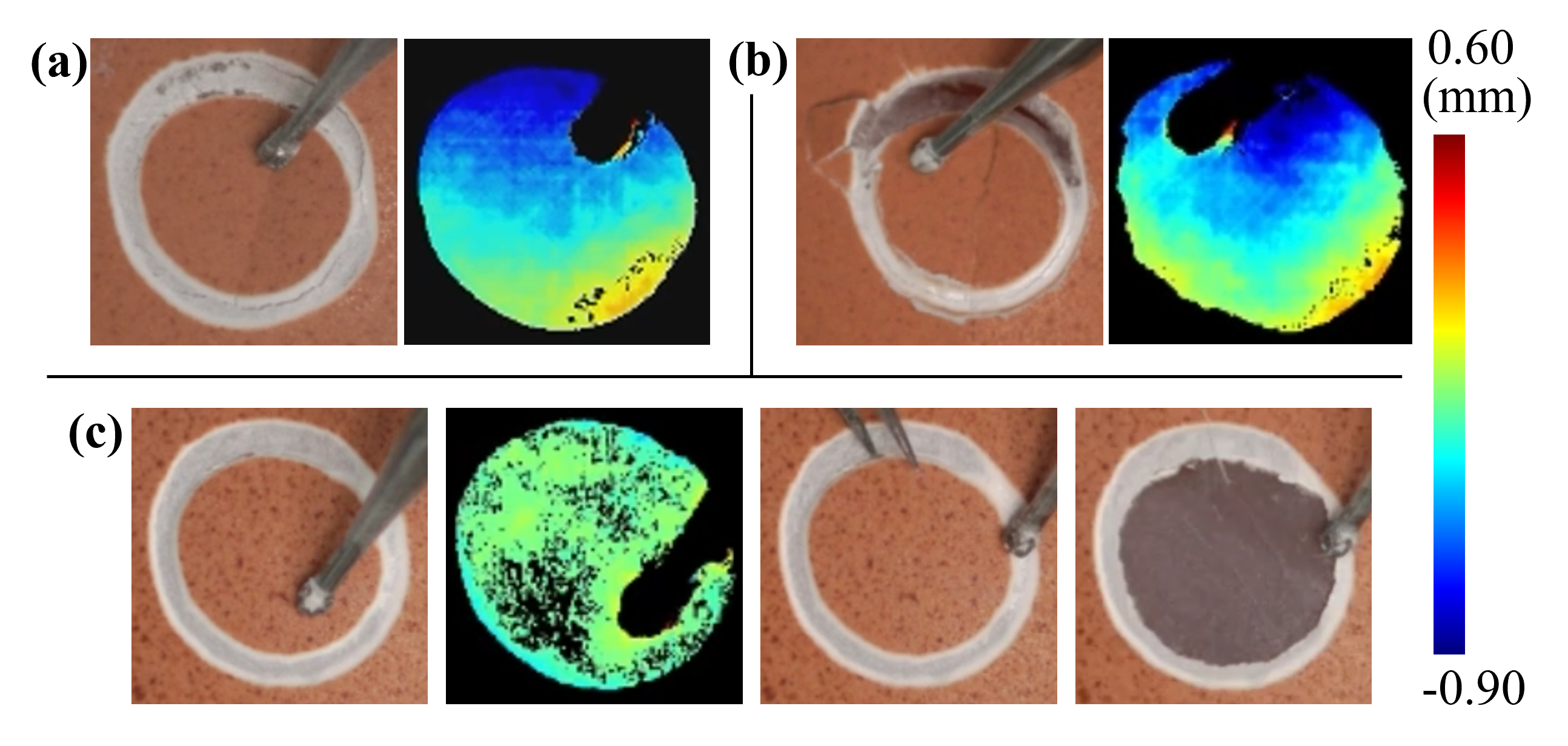}
\caption{Examples of the experiment results. (a) and (b) showcase Case 1 and Case 2, respectively, derived from Trial 10 and Trial 1 (see Table~\ref{table:1}). The left RGB images demonstrate the result of success or failure, while the right images illustrate detachability via depth maps. (c) showcases Case 3, derived from Trial 6. The first and second images on the left indicate that the bone flap was in a non-detachable state when drilling stopped, while the third and fourth images show the attempt to \textbf{forcibly} remove the bone flap using tweezers manually and the successful result.}
\label{fig:measurement}
\end{figure}

Combining the two metrics, the results can be summarized into four cases: Case 1: successful and detachable; Case 2: failed but detachable; Case 3: successful but non-detachable; and Case 4: failed and non-detachable. Case 1 is the ideal case, accounting for 66.7\% of this experiment (previous system: 40\%). Trial 10 is an example of Case 1, as shown in Fig.~\ref{fig:measurement}-(a). Case 2 is caused by over-drilling, leading to significant membrane damage, and it represents the only failure trial (trial 1, shown in Fig.~\ref{fig:measurement}-(b)), accounting for 8.3\% of this experiment (previous system: 10\%). Case 3 occurs when the experiment ends with the bone flap estimated as non-detachable, but it can be forcibly removed by human. This case accounts for 25\% of this experiment (previous system: 40\%), with Trial 6 being one example, as shown in Fig.~\ref{fig:measurement}-(c). Case 4 is not included in Fig.~\ref{fig:measurement} nor the experimental results. It occurs when the drilling completion level is too low, preventing the bone flap from being detached or forcibly removed, which did not occur in this experiment but accounted for 10\% in the previous system. 

Therefore, it can be concluded from the result that the proposed object state estimation method effectively improves the performance of the ARDS in eggshell drilling, both in terms of success and detachable ratio, by increasing the proportion of Case 1 and reducing the proportion of Case 3 and 4. It is also important to note that in Trial 7, the drilling process automatically stopped at an early stage by the exception handling (described in Section~\ref{subsec:exception}), which demonstrates the proposed system's accuracy in detecting deflection and highlights its robustness in handling intraoperative exceptions. 

However, we also observe that a longer drilling time is required with the improved system. This increase is due to the palpating phases, which account for 28.1\% of the total drilling time. This is a limitation of the current system, where further fine-tuning parameters are necessary to reduce the proportion of palpation time and the total drilling time.




Other limitations include that the current system cannot resolve Case 2, which is caused by over-drilling due to recognition errors from the Completion Level Recognition module. Additionally, the recognition errors from the Completion Level Recognition module also contribute to Case 3, where the latest generated trajectory used in the Repeat Drilling phase is insufficient to make the bone flap detachable, which contradicts our assumption (Section~\ref{subsec:repeatdrilling}). Case 3 is also one of the cases we aim to avoid, whose proportion has decreased compared to previous experiments, though 25\% is still relatively high and needs to be improved.



\section{Discussion}

Our study introduced a novel object state estimation method through robotic active interaction, advancing our previous efforts on autonomous micro medical drilling. This method was rigorously tested through robotic experiments to validate its effectiveness and performance within our improved ARDS. At the same time, our method also lays a good foundation for possible subsequent research on autonomous robotic removal of bone flaps of biological specimens.

When removing a piece or flap from a specimen, it is intuitive to observe the movement of the flap by interaction. A minute deflection of the flap on a microscopic scale is difficult to observe. Instead, people will watch the change of light reflection on the object's surface during the interaction, indicating the object's state. However, with a high-resolution stereo camera system, the deflection can be directly detected, which greatly facilitates the estimation of object state in the manipulation. Moreover, this method provides an alternative means of tactile sensing where the specimen is deformable and too small to apply a tactile sensor. 

Our ultimate goal is to adapt this methodology for craniotomies on mice skulls. For practical reasons, our experiments utilized eggs as a surrogate, in line with the rationale and assumptions outlined in our earlier work \cite{zhao2023}. The Segmentation \& Masking method in the Deflection Detector is implemented by an HSV thresholding technique, which works for eggshells but lacks effectiveness for a real cranial surface, where a learning-based segmentation method will be needed. 

Despite the state estimation in this research being solely based on deflection, it is sufficient for eggshell application because the states are limited to two situations due to the rigidity of the specimen. The bone flap is recognized as detachable once it is movable. Otherwise, it is regarded as non-detachable. Force information ensures safety during the palpation so that it will not break the object in case it is still non-detachable. For a more flexible biological specimen like a mouse skull, it is necessary to consider the change in stiffness as a metric to assess the object's state and estimate the completion level. Incorporating force information for estimating the stiffness to further infer the state will be one of the directions in our future work.



\section{Conclusion}

This study proposes an object state estimation method via active robotic interaction, which is then integrated into the ARDS to improve the automation workflow of craniotomy procedures. Our experiments, conducted with this improved system on eggshell drilling, have achieved a 91.7\% successful ratio and 75\% detachable ratio in operational terms. This foundational work lays the groundwork for future applications in more complex surgical contexts, such as craniotomy procedures on mice and using force information in state estimation on more flexible biological specimens.


 
\bibliographystyle{IEEEtran}
\bibliography{references}

\newpage

\vfill

\end{document}